\documentclass[english,a4paper,11pt]{article}\PassOptionsToPackage{pdfpagelabels=false}{hyperref} 
\usepackage{fullpage}
\usepackage{amsxtra, amsfonts, amssymb, amstext}
\usepackage{amsthm}
\usepackage{booktabs}
\usepackage{longtable}
\usepackage{nicefrac}
\usepackage{xspace}
\usepackage[noadjust]{cite}
\usepackage{url}
\usepackage{graphics,color}
\usepackage[colorlinks]{hyperref}
\definecolor{linkblue}{rgb}{0.1,0.1,0.8}
\hypersetup{colorlinks=true,linkcolor=linkblue,filecolor=linkblue,urlcolor=linkblue,citecolor=linkblue}
\usepackage[algo2e,ruled,vlined,linesnumbered]{algorithm2e}
\newcommand{\assign}{\leftarrow}
\usepackage{graphicx}
\usepackage{wrapfig}
\usepackage{caption}
\usepackage{subcaption}
\newtheorem{theorem}{Theorem}
\newtheorem{lemma}[theorem]{Lemma}

\allowdisplaybreaks[1]
\newcommand{\N}{\mathbb{N}}
\newcommand{\R}{\mathbb{R}}
\DeclareMathOperator{\mut}{mut}

\DeclareMathOperator{\opt}{opt}

\newcommand{\E}{\mathbb{E}}
\renewcommand{\Pr}{\mathbb{P}}

\DeclareMathOperator{\Bin}{Bin}
\newcommand{\onemax}{\textsc{OneMax}\xspace}
\newcommand{\OM}{\textsc{Om}\xspace}
\newcommand{\leadingones}{\textsc{LeadingOnes}\xspace}
\newcommand{\LO}{\textsc{Lo}\xspace}
\newcommand{\oea}{$(1 + 1)$~EA\xspace}
\DeclareMathOperator{\RLS}{RLS}
\newcommand{\RLSopt}{RLS$_{\text{opt}}$\xspace}
\newcommand{\oeares}{$(1 + 1)$~EA$_{>0}$\xspace}
\newcommand{\oeaalpha}{$(1 + 1)$~EA$_{\alpha}$\xspace}
\newcommand{\oeaalphatwo}{$(1 + 1)$~EA$_{\alpha}$}
\title{On the Effectiveness of Simple Success-Based Parameter Selection Mechanisms for Two Classical Discrete Black-Box Optimization Benchmark Problems}
\author{Carola Doerr$^1$ and Markus Wagner$^2$}
\date{
$^1$Sorbonne Universit\'e, CNRS, Laboratoire d'Informatique de Paris 6, LIP6, 75005 Paris, France\\
$^2$Optimisation and Logistics, The University of Adelaide, Adelaide, SA 5005, Australia
}
\begin{document}
\maketitle 

{\sloppy
\begin{abstract}
Despite significant empirical and theoretically supported evidence that non-static parameter choices can be strongly beneficial in evolutionary computation, the question \emph{how} to best adjust parameter values plays only a marginal role in contemporary research on discrete black-box optimization. This has led to the unsatisfactory situation in which feedback-free parameter selection rules such as the cooling schedule of Simulated Annealing are predominant in state-of-the-art heuristics, while, at the same time, we understand very well that such time-dependent selection rules can only perform worse than adjustment rules that \emph{do} take into account the evolution of the optimization process. A number of adaptive and self-adaptive parameter control strategies have been proposed in the literature, but did not (yet) make their way to a broader public. A key obstacle seems to lie in their rather complex update rules. 

The purpose of our work is to demonstrate that high-performing online parameter selection rules do not have to be very complicated. More precisely, we experiment with a multiplicative, comparison-based update rule to adjust the mutation probability of a (1+1)~Evolutionary Algorithm. We show that this simple self-adjusting rule outperforms the best static unary unbiased black-box algorithm on LeadingOnes, achieving an almost optimal speedup of about~$18\%$.
\end{abstract}

\section{{Introduction}\label{sec:intro}}

One of the best known randomized black-box optimization techniques is Simulated Annealing~\cite{SA83}. Simulated Annealing builds on the \emph{Metropolis} heuristic, a randomized search method that aims to overcome the risk of getting stuck in local optima by allowing the search to continue in points that are worse than the current best solution. The probability to ``accept'' such an inferior solution depends on the absolute difference in function values and upon a parameter $T$, which is often referred to as the \emph{temperature} of the system. The improvement of Simulated Annealing over the Metropolis algorithm is a non-static choice of this temperature $T$. By decreasing $T$ over time, the search algorithm converges from an exploratory behavior to a more and more greedy hill-climber, which exploits the good regions identified during the earlier phases. Numerous successful applications and more than 42,000 citations of~\cite{SA83} witness that this idea to \emph{control} the temperature during the optimization process can have an impressive impact on the performance of the Metropolis algorithm. 

It is today quite well understood that non-static parameter choices can be quite beneficial also for evolutionary computation (EC) methods, cf.~\cite{KarafotiasHE15,DoerrGWY17,Carola17}. The question \emph{how} to adjust the parameters, however, is largely open. Already the selection of suitable \emph{static} parameter values is a very complex problem that has given rise to a number of sophisticated parameter tuning techniques; it is not for nothing that the parameter selection problem is coined the ``Achilles' heel of evolutionary computation'' in~\cite{FialhoCSS10}. Finding methods that automatically detect \emph{and track} optimal parameter values over all stages of the optimization process are the long-term vision of research on \emph{parameter control} techniques. 

Adaptive parameter choices are indispensable in continuous optimization, and therefore used in most state-of-the-art heuristics. In \emph{discrete domains}, however, the situation is quite different. A number of different parameter control techniques have been experimented with in the literature, but so far none of them has been able to establish itself as a standard technique. Quite the contrary, the vast majority of research papers on discrete black-box optimization problems use static parameters values~\cite{KarafotiasHE15}. In light of the success story of Simulated Annealing, this situation is quite unsatisfactory. 

A potential reason for the discrepancy between the commonly acknowledged benefits of non-static parameter selection schemes and its low degree of utilization may be the complexity of the parameter control techniques that predominate in the EC literature: these are often based on self-adaptation, hyper-heuristics, or principles from machine learning~\cite{DangL16PPSN,DoerrDY16PPSN,KarafotiasHE15,LissovoiOW17}. Such (more or less) sophisticated techniques are in sharp contrast to the rather simple cooling schedule of Simulated Annealing, which updates the temperature based on the time elapsed so far. Of course, we easily convince ourselves that such a feedback-free update rule cannot be optimal, and that more efficient parameter control techniques take into account the behavior of the optimization process, such as, for example, the rate of success, the magnitude of improvement achieved within a given time-window, or the diversity of the population. 

\subsection{Our Results}

We analyze in this work one of the simplest ways to incorporate feedback from the optimization process into the selection of suitable parameter values: a \emph{success-based multiplicative update rule.} To abstract away potential inter-dependencies between multiple parameters, we concentrate on the control of a single parameter, the mutation rate of a (1+1) Evolutionary Algorithm (EA) that does not evaluate offspring that are identical to their parents (cf. Section~\ref{sec:prelims} for details and motivation). The update of the mutation rate depends only on whether or not the parent individual is replaced by the offspring. When no improvement is found, the mutation rate $p$ is decreased to $bp$ (with the idea to be more conservative), and it is increased to $Ap$ otherwise (motivated by the desire to make even more progress by searching at a larger distance). $A$ and $b$ are \emph{hyper-parameters} that satisfy $A>1$ and $0<b<1$. This multiplicative control technique is entirely \emph{comparison-based}, a highly desirable property for online parameter selection mechanisms~\cite{FialhoCSS10}. 

We investigate the performance of the adaptive \oea variant, the \oeaalpha, on \onemax and on \leadingones. The experiments on \onemax confirm that, for a very broad range of update strengths, the \oeaalpha is capable of identifying optimal parameter values ``on-the-fly''. This leads to average optimization times that are very close to being optimal among all unary unbiased black-box algorithms.\footnote{We recall that unary unbiased algorithms are those that sample all search points uniformly at random from the whole search space or from unbiased distributions that depend on exactly one previously evaluated search point. A distribution is \emph{unbiased} if it does not discriminate between bit positions, not between bit values~\cite{LehreW12}.} 

For \onemax in reasonable problem dimensions $n \le 20,000$, the relative advantage of non-static parameter choices is only around $2\%$, and thus not very pronounced. The main task of the \oeaalpha is therefore to \emph{identify} good (i.e., in the context of \onemax, low) mutation rates, and to not get distracted by a potential desire to greedily increase the mutation rate during the optimization. 

To investigate the ability of the \oeaalpha to not only identify but to also track optimal parameter values that change quite drastically during the optimization process, we also regard its performance on the classic \leadingones benchmark. For \leadingones, the optimal number of bits to flip depends on the fitness of the current-best individual. It is $n$ for $x$ with $\LO(x)=0$, $n/2$ for search points of fitness $1$, and decreases to $1$ for search points of \LO-values $\ge n/2$, cf. Lemma~\ref{lem:koptLO}. Our empirical results confirm that, again for a broad range of hyper-parameters, the \oeaalpha is able to find and to \emph{track} these optimal mutation rates. About $62\%$ of all $450$ tested configurations with $1<A\le 2.5$ and $0.4<b<1$ and around $39\%$ of all $2,450$ configurations with $1<A\le 6$ and $0<b<1$ outperform Randomized Local Search, the best unary unbiased black-box algorithm with static mutation rates, by at least $10\%$. Some configurations achieve an almost optimal advantage of around $18\%$.  

\textbf{Disclaimer.} A common critique of fundamental research on \onemax and similar benchmark problems is that such ``sterile'' environments are not very representative for typical applications of EAs. Based on the results presented in this work, we, of course, do not know to what extend the advantages of the success-based multiplicative update rule applies to more complex optimization problems. It has been argued, however, that results for \onemax can serve, at the very least, to verify if some important properties of parameter control techniques are satisfied~\cite{Thierens09,FialhoCSS09}. In this sense, our results can be seen as a ``proof of concept'' for the largely unexplored potential of parameter control. 
We also wish to point out that multiplicative update rules have of course been studied much prior to this work, e.g.,~\cite{LassigS11,DoerrD15self,DoerrDK16PPSN,JansenJW05}. The appealing aspect of our study lies in the simplicity of the algorithm and problems, which help to nicely illustrate the working principles of this promising parameter update rule. We hope that the convincing and detailed empirical evaluation serves as a motivation to experiment with parameter control techniques. The case of Simulated Annealing has shown that paradigm changes are possible, and we feel that it is time for EC methods to overcome static and feedback-free parameter selection mechanisms.  

\section{Algorithms and Benchmarks}\label{sec:prelims}

Our study aims at quantifying the positive effects of online parameter selection. To remove any unwanted side effects, we therefore remove the population size and selective pressure as parameters, and concentrate on adapting the mutation rate in the two classical black-box optimization algorithms Randomized Local Search (RLS) and the (1+1) Evolutionary Algorithm (EA). We use this section to describe the algorithms and benchmark problems studied in our work. For the hasty reader, we wish to point out that we regard a variant of the \oea in which we ensure that an offspring does not equal its direct parent (i.e., we do not allow offspring to be copies of their parent's genotype). 

\textbf{Notation.} The description of the algorithms assumes the \emph{maximization} of a pseudo-Boolean function $f:\{0,1\}^n \to \R$ as optimization task. By $[n]$ we abbreviate the set $\{1,2,\ldots,n\}$, and we let $[0..r]:=\{0\} \cup [r]$.

\subsection{RLS and the Resampling (1+1) EA}
RLS and the \oea are (1+1) schemes. That is, they always maintain one previously queried solution $x$, sample from it exactly one offspring $y$, and use elitist selection; that is, $y$ replaces $x$ if any only if $f(y)\ge f(x)$. The difference of RLS and the \oea lies in the generation of $y$. While RLS creates the \emph{offspring} $y$ by flipping exactly one bit in $x$ that is chosen uniformly at random, the \oea creates $y$ by \emph{standard bit mutation}. That is, $y=(y_1,\ldots,y_n)$ is selected by first copying $x$ and then flipping each bit with some positive probability $p$, independently of all other bits. The parameter $p$ is referred to as the \emph{mutation probability} or \emph{mutation rate}. A standard choice for $p$ is $1/n$, which results in an expected number of one bit flip. Put differently, an average iteration of the \oea with mutation probability $1/n$ behaves like an RLS iteration, with the difference that standard bit mutation is a \emph{global variation operator:} in every iteration, every search point $x \in \{0,1\}^n$ has a positive probability to be sampled. Our main interest is in studying adaptive choices of $p$, but before we discuss our adaptation rules, we recall one important observation about standard bit mutation. 

It is not very difficult to see that standard bit mutation can be identically defined by first choosing a \emph{step size} (aka \emph{mutation strength}) $\ell$ from the binomial distribution $\Bin(n,p)$ and then applying the $\mut_{\ell}$ variation operator (Algorithm~\ref{alg:mut}), which samples $\ell$ different indices uniformly at random and creates an offspring $y$ from $x$ by flipping the bits in these $\ell$ positions and copying the values from $x$ elsewhere. Note that with this description, RLS is the algorithm that uses in each iteration the operator $\mut_{1}$, i.e., it chooses $\ell=1$ deterministically.

\begin{algorithm2e}	\textbf{Input:} $x \in \{0,1\}^n$, $\ell \in \N$\;
		\label{line:elloea}Select $\ell$ different positions $i_1,\ldots,i_{\ell} \in [n]$ u.a.r.\;
	  $y \assign x$\;
		\lFor{$j=1,...,\ell$}{$y_{i_j}\assign 1-x_{i_j}$}
\caption{$\mut_{\ell}$ chooses $\ell$ different positions and flips the entries in these positions.}
\label{alg:mut}
\end{algorithm2e}

It was noted in~\cite{JansenZ14,CarvalhoD17,HemertB02} that the literate implementation of standard bit mutation is inadequate for most practical purposes, since the probability that an offspring is identical to its direct parent equals $\Bin(n,p)(0)=(1-p)^n$, which for $p=1/n$ converges very quickly to $1/e \approx 36.8\%$. Unless we are dealing with very noisy function evaluations, which is not the situation regarded here, such offspring do not advance the optimization process, as they do not carry any new information about the problem instance. Any reasonable implementation of the \oea would therefore not evaluate such offspring. An efficient implementation of the \oea would therefore avoid to generate such offspring in the first place. This is easily possible, as all we need to do is to re-sample the step size $\ell$ from $\Bin(n,p)$ until we get a non-zero value. This is identical to sampling $\ell$ from the conditional distribution $\Bin_{>0}(n,p)$, which assigns probability $0$ to the step size $0$ and probability $\Bin_{>0}(n,p)(k)=\binom{n}{k}p^p(1-p)^{n-k}/(1-(1-p)^n)$ to any positive step size $k>0$. According to~\cite{CarvalhoD17}, this resampling strategy seems to be a very common implementation of standard bit mutation in plus strategies like the \oea. To distinguish this interpretation of the \oea from the one classically regarded in the theory of evolutionary computation literature, it is named \oeares in~\cite{CarvalhoD17}.

\subsection{Self-Adaptive Mutation Rates}

As mentioned in the introduction, our main interest is in the study of performance gains that can be achieved by a non-static choice of the mutation rate $p$. To this end, we investigate the following simple update rule. If an iteration was successful, i.e., if it produced an offspring $y$ that replaces $x$, we increase the mutation probability $p$ by a constant multiplicative factor $A\ge 1$. That is, we replace $p$ by $Ap$ if $f(y)\ge f(x)$ holds. If, on the other hand, $y$ is discarded ($f(y) < f(x)$), we decrease $p$ to $bp$, where $b \le 1$ is again some fixed constant. We cap the value of $p$ to ensure that it is always greater than $1/n^2$ and at most $1/2$. The initial value of $p$ is set to $p_0$, for some constant $p_0>0$. This gives the \oeaalphatwo$(A,b,p_0)$, which we summarize in Algorithm~\ref{alg:oeaalpha}. Note that the \oeaalphatwo$(A=1,b=1,p_0)$ is the \oeares with static mutation rate $p=p_0$. 

 \begin{algorithm2e}	\textbf{Initialization:} 
	Sample $x \in \{0,1\}^{n}$ uniformly at random and compute $f(x)$\;
	Set $p=p_0$\;
  \textbf{Optimization:}
	\For{$t=1,2,3,\ldots$}{
		\label{line:ellres}Sample $\ell$ from $\Bin_{>0}(n,p)$\;
		$y \assign \mut_{\ell}(x)$\;
		evaluate $f(y)$\;
		\eIf{$f(y)\geq f(x)$}
			{$x \assign y$ and $p \assign \min\{ A \cdot p, 1/2\}$}
			{$p \assign \max\{b \cdot p, 1/n^2\}$}	
}\caption{The \oeaalpha with update strengths $A$ and $b$ and initial mutation probability $p_0\in [1/n^2,1/2]$ for the maximization of a pseudo-Boolean function $f:\{0,1\}^n \rightarrow \R$}
\label{alg:oeaalpha}
\end{algorithm2e}

In our experiments we will compare the performance of the \oeaalpha with a variant of RLS that uses a non-static choice of the step size. This variant will be described in Section~\ref{sec:theory}.

\subsection{OneMax and LeadingOnes}
As benchmark problems, we select \onemax and \leadingones, since for these two problems we understand quite well how the optimal mutation strengths depend on the state of the optimization process, so that we have a solid baseline against to which we can compare the performance of the \oeaalpha. \onemax and \leadingones are both problems with a unique global optimum $z \in \{0,1\}^n$. 

\textbf{OneMax.} For every \emph{target string} $z \in \{0,1\}^n$, the \onemax function $\OM_z$ assigns to each search point $x \in \{0,1\}^n$ the number of positions in which $x$ and $z$ agree, i.e., $\OM_z(x):=|\{i \in [n] \mid x_i=z_i \}|$. Maximizing $\OM_z$ corresponds to minimizing the Hamming distance between $x$ and $z$. 

It is well known that every mutation-based algorithm (in the unary unbiased sense promoted in~\cite{LehreW12}) needs $\Omega(n \log n)$ function evaluations, on average, to optimize \onemax. In this asymptotic sense, all RLS and all static \oea variants considered in this work are optimal, since they are all unary unbiased algorithms and they all achieve a $\Theta(n \log n)$ expected optimization time. We will nevertheless see that the actual (i.e., non-asymptotic) running time can differ substantially for the different algorithms. We will discuss more precise running time statements in Section~\ref{sec:theory} below.  

\textbf{LeadingOnes} is the problem of optimizing an unknown function of the type $\LO_{z,\sigma}:\{0,1\}^n \to \R, x \mapsto \max \{ i \in [0..n] \mid \forall j \in [i]: x_{\sigma(i)}=z_{\sigma(i)} \}$, where $z$ is an unknown length-$n$ bit string and $\sigma:[n] \to [n]$ an unknown permutation (one-to-one map) of the positions. That is, $\LO_{z,\sigma}(x)$ is the length of the longest common prefix between $z$ and $x$ in the order determined by $\sigma$. 

Every (1+1) elitist~\cite{DoerrL16} and every unary unbiased~\cite{LehreW12} black-box algorithm needs $\Omega(n^2)$ function evaluations, on average, to optimize \leadingones. This bound is matched by RLS and the \oea, as we shall discuss in the next section. 

\section{Theoretical Performance Limits}\label{sec:theory}

To establish bounds against which we can compare the \oeaalpha, we now take a closer look at the best possible performance that any mutation-based algorithm can achieve on \onemax and \leadingones. In both cases, this performance is obtained by a variant of RLS that replaces the static choice $\ell=1$ classically used by RLS by a fitness-dependent step size $\ell$. More precisely, it is known that, for \onemax, such an RLS variant has a performance that can not be worse than an optimal unary unbiased black-box algorithm by more than an additive $o(n)$ term~\cite{DoerrDY16}. For \leadingones a similar statement can be derived from the methods introduced in~\cite{BottcherDN10,DoerrDY16}, cf. Section~\ref{sec:theoLO}.

\subsection{{OneMax}\label{sec:theoOM}}

As mentioned above and summarized in~\cite{DoerrDY16}, we know quite well how RLS and the \oea perform on the \onemax problem. From the known bounds, we can compute theoretical performance limits of the \oeaalpha. This is the focus of this section.

For \textbf{static parameter values,} i.e., for arbitrary $p_0 \in [1/n^2,1/2]$ and $A=b=1$, the expected optimization time of the \oeaalpha on \onemax cannot be better than that of RLS, which is equal to $n \ln(n) + (\gamma - \ln(2)) + o(1) \approx n \ln n - 0.1159n$~\cite{DoerrD16impact} ($\gamma = 0.5772\dots$ denotes the Euler-Mascheroni constant). Likewise, for \textbf{adaptive parameter values} (i.e., for arbitrary values of $A$, $b$, and $p_0$) the expected optimization time of the \oeaalpha is bounded from below by the performance of a best unary unbiased black-box algorithm, which satisfies $n \ln(n) - \alpha n \pm o(n)$ for a constant $\alpha$ that is between $0.2539$ and $0.2665$~\cite{DoerrDY16}. 

The above-mentioned values are asymptotically optimal running times. In order to obtain \emph{absolute performance limits for concrete problem dimensions,} we regard the \emph{drift-maximizing} RLS variant studied in~\cite{DoerrDY16}. Although it cannot be formally proven that this algorithm is indeed optimal, the result in~\cite{DoerrDY16} states that its performance cannot be much worse than that of the best possible unary unbiased (i.e., mutation-based) black-box algorithm. We even conjecture that the drift-maximizing RLS, which we call $\RLS_{\text{opt,OM}}$, \emph{is} indeed optimal within this class. 

$\RLS_{\text{opt,OM}}$ is the RLS variant that modifies the best-so-far solution $x$ by applying to it the variation operator $\mut_{\ell}$ for a value of $\ell$ that maximizes the expected progress that can be obtained in one iteration. This expected progress is often referred to as \emph{drift}, hence the name ``drift maximizer''. It is not difficult to see that the expected progress $\E\Big[\max\{\OM(\mut_{\ell}(x))-\OM(x),0\}\Big]$ of $\mut_{\ell}$ applied to $x$ equals 
\begin{align}\label{eq:driftOM}
	\sum_{i=\lceil \ell/2 \rceil}^{\ell}
	\frac{\binom{n-\OM(x)}{i} \binom{\OM(x)}{\ell-i}\left(2i-\ell \right)}{\binom{n}{\ell}}.
\end{align}
This expression depends only on the problem dimension $n$ and the function value $\OM(x)$, but not on the structure of the search point~$x$. For every $n$ and every possible function value $f \in [0..n]$, we can therefore abbreviate the progress-maximizing choice of $\ell$ by $k_{\text{opt,OM}}(n,f)$. With this abbreviation, $\RLS_{\text{opt,OM}}$ is Algorithm~\ref{alg:RLSoptOM}. 

As proven in~\cite{DoerrDY16}, the value of $k_{\text{opt,OM}}(n,f)$ equals $1$ whenever $f \ge 2n/3$. For general $f$, however, we do not have a simple to evaluate closed form expression to describe $k_{\text{opt,OM}}(n,f)$. For this reason an approximation of $k_{\text{opt,OM}}(n,f)$ is used in~\cite{DoerrDY16}. Since here in this work we are not interested in asymptotic bounds, but rather absolute values for concrete problem dimensions, we do not need to approximate $k_{\text{opt,OM}}(n,f)$ but can work with the exact drift maximizing choice.  Using these values, we can evaluate the expected performance of $\RLS_{\text{opt,OM}}$ empirically. This is our approach for the results presented in Sections~\ref{sec:grid} and~\ref{sec:experimentsOM}. These empirical averages are quite close to the above-mentioned asymptotic lower bound presented in~\cite{DoerrDY16}. An alternative way would be to apply a variable drift theorem to the point-wise drift, i.e., expression~\eqref{eq:driftOM} evaluated for $\ell=k_{\text{opt,OM}}(n,\OM(x))$. 
 \begin{algorithm2e}	Sample $x \in \{0,1\}^{n}$ uniformly at random and compute $\OM(x)$\;
	\For{$t=1,2,3,\ldots$}{
		\label{line:RLSoptOMmut} $\ell \assign k_{\text{opt,OM}}(n,\OM(x))$\;
		$y \assign \mut_{\ell}(x)$\;
		\lIf{$f(y)\geq f(x)$}
			{$x \assign y$}	
}\caption{The drift-maximizing algorithm $\RLS_{\text{opt,OM}}$}
\label{alg:RLSoptOM}
\end{algorithm2e}

\subsection{{LeadingOnes}\label{sec:theoLO}}

For \leadingones the known theoretical bounds are as follows. RLS needs $1+n^2/2$ function evaluations, on average, for its optimization. This bound is also a lower bound for the \oeaalpha with \textbf{static parameter choices}; i.e., for arbitrary $p_0 \in [1/n^2,1/2]$ and $A=b=1$. For the classical \oea, which samples $\ell$ from the unconditional binomial distribution $\Bin(n,p)$ (and may therefore sample $\ell=0$), the best static choice is $p\approx 1.59n$, which gives an expected optimization time of about $0.77 n^2$, while for the \oeares it holds that the smaller the mutation rate, the better performance we obtain~\cite{JansenZ11}. More precisely, it holds that the expected running time of the \oeares converges to $1+n^2/2$ when the mutation rate $p$ converges to zero.

In~\cite{BottcherDN10}, also optimal \textbf{adaptive mutation rates} have been computed for the classical \oea. It is shown there that the \oea using at each point $x$ the mutation rate $(n-\LO(x))/n$ has an expected running time on \leadingones of $0.68 n^2 \pm O(n)$. This is optimal among all \oea variants that are charged for 0-bit flip iterations. 

The adaptive \oea variant from~\cite{BottcherDN10} clearly looses performance for iterations in which the offspring equals its parent. It is therefore natural to ask for the best performance that a---possibly adaptive---unary unbiased black-box algorithm can achieve. As far as we know, such a best-possible mutation-based algorithm has not been explicitly reported in the literature. It turns out, however, that we can generalize mathematical statements proven in~\cite{BottcherDN10} and~\cite{DoerrDY16} to design such an optimal unary unbiased black-box algorithm for \leadingones.

Before describing this algorithm in detail, we note that already allowing 1- and 2-bitflips (i.e., $\mut_{\ell}$ with $\ell=1$ and $\ell=2$) decreases the optimal $1+n^2/2$ expected optimization time of static unary unbiased algorithms to about $0.4233 n^2$~\cite{LissovoiOW17} (the fact that $\mut_{\ell}$ is defined slightly different in~\cite{LissovoiOW17} has a negligible impact on this result). This running time can be further reduced by allowing larger step sizes. We investigate the limits of this approach in the remainder of this section. 

To compute the progress-maximizing variant of RLS, we could, similarly to the \onemax case, compute the expected progress of $\mut_{\ell}$ when applied to a search point $x$. For \leadingones, however, it suffices to maximize the \emph{probability} of making progress~\cite{BottcherDN10}, which is, in general, much easier than computing the expected progress. When applied to a search point $x$, the offspring $y$ created from $x$ by flipping $\ell$ bits satisfies $\LO(y)>\LO(x)$ if and only if the $(\LO(x)+1)$-st bit is flipped but none of the first $\LO(x)$ bits. The probability of this event, for uniformly chosen bit flips, equals 
\begin{align}\label{eq:probLO}
	\Pr\Big[\LO(\mut_{\ell}(x)) > \LO(x)\Big]
& = 
	\frac{\ell}{n-\LO(x)}\binom{n-\LO(x)}{\ell}/\binom{n}{\ell}\\
& \nonumber = 
	\binom{n-\LO(x)-1}{\ell-1}/\binom{n}{\ell}.
\end{align} 
In line with the notation used for the \onemax case, we abbreviate the value $\ell$ that maximizes expression~\eqref{eq:probLO} by $k_{\text{opt,LO}}(n,\LO(x))$. The following lemma seems to be well known in the theory of evolutionary computation community, but, as far as we know, it has not been mentioned explicitly. 

\begin{lemma}[$k_{\text{opt,LO}}(n,\LO(x))$] \label{lem:koptLO}
For all $n \in \N$ and for all $x \in \{0,1\}^n$ it holds that $$k_{\text{opt,LO}}(n,\LO(x)) = \lfloor n/(\LO(x)+1) \rfloor.$$
\end{lemma}

With these values, we can study the expected running time of $\RLS_{\text{opt,LO}}$, which is Algorithm~\ref{alg:RLSoptOM} with line~\ref{line:RLSoptOMmut} replaced by ``$\ell \assign k_{\text{opt,LO}}(n,\LO(x))$''. Combining Lemma~\ref{lem:koptLO} with the characterization of unary unbiased mutation operators provided in~\cite[Lemma~1]{DoerrDY16} and an extension of the results proven in~\cite{BottcherDN10} to unary unbiased black-box algorithms, it is not difficult to show the following theorem, which, intuitively speaking, states that $\RLS_{\text{opt,LO}}$ is optimal among all mutation-based black-box algorithms for \leadingones. Furthermore, this optimality does not only apply to the overall optimization time, but also to all intermediate target values. We thus obtain the following statement about the \emph{fixed-target performance} of $\RLS_{\text{opt,LO}}$.

\begin{theorem}\label{thm:LOopt}
For \leadingones, the expected number of function evaluations needed by $\RLS_{\text{opt,LO}}$ to obtain a search point of function value at least $i$ equals 
\begin{align*}T(\RLS_{\text{opt,LO}},\LO,i)
& :=1+ \frac{1}{2}\sum_{j=0}^{i-1}{\binom{n}{k_{\text{opt,LO}}(n,j)}/\binom{n-j-1}{k_{\text{opt,LO}}(n,j)-1}}\\
&\nonumber 
= 1+ \frac{1}{2}\sum_{j=0}^{i-1}{\binom{n}{\lfloor n/(j+1) \rfloor}/\binom{n-j-1}{\lfloor n/(j+1) \rfloor-1}}.
\end{align*}
For all $n$ and all $i$ this performance is optimal among all unary unbiased black-box algorithms. That is, for any unary unbiased black-box algorithm $A$ the expected time needed by $A$ to reach a search point of \leadingones value $\ge i$ is at least as large as that of $\RLS_{\text{opt,LO}}$.
\end{theorem}

We did not find an easy to evaluate closed form for $T(\RLS_{\text{opt,LO}},\LO,i)$. We can nevertheless evaluate this sum numerically, and obtain that for $n \rightarrow \infty$ the expected optimization time of $\RLS_{\text{opt,LO}}$ seems to converge to around $0.388... n^2$, cf. Table~\ref{tab:LOn}. For $n=10,000$ the value is still around $0.3884 n^2$. 

\section{Grid search}\label{sec:grid}

\begin{figure*}
        \centering
			
        \begin{minipage}[t]{0.33\textwidth}
                \includegraphics[trim={12mm 0 2mm 0},width=\textwidth]{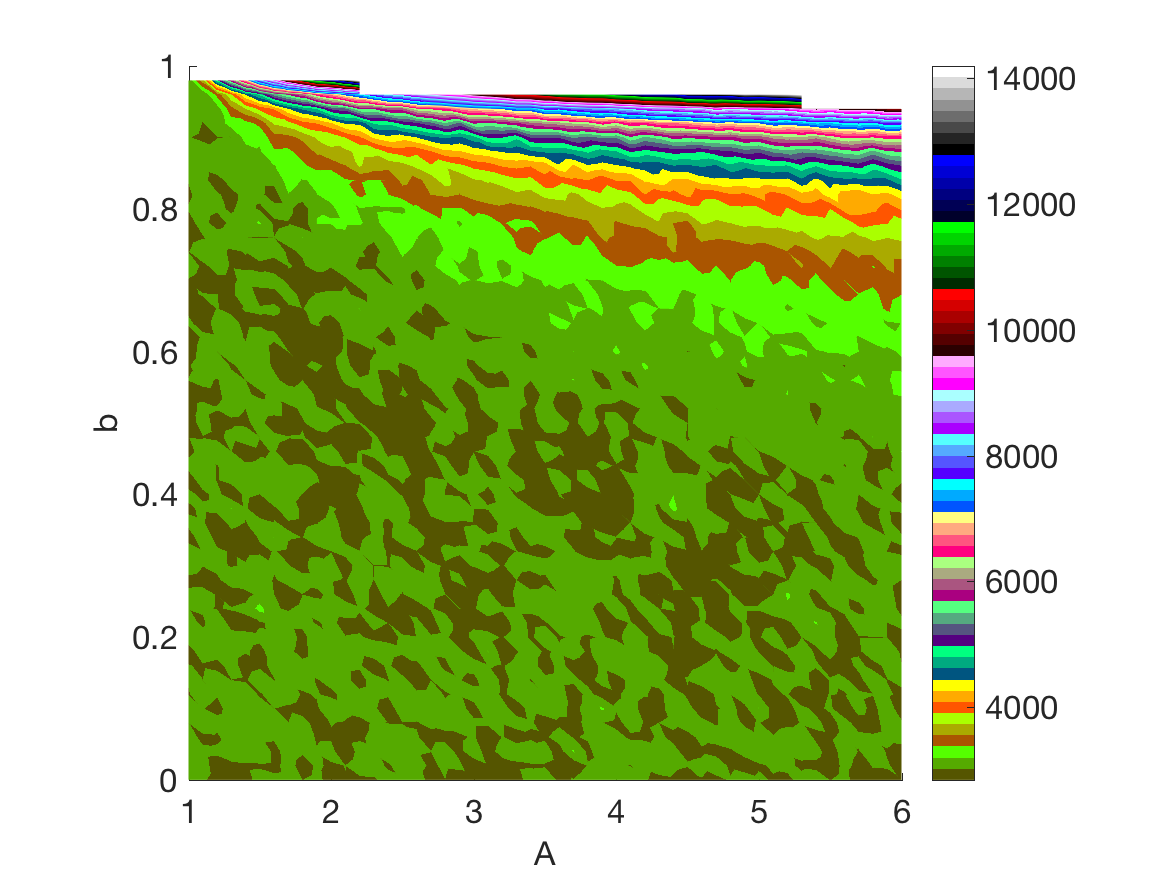}
                \subcaption{\onemax with $n=500$}\label{fig:OM500}
        \end{minipage}\hfill
				\begin{minipage}[t]{0.33\textwidth}
                \includegraphics[trim={12mm 0 2mm 0},width=\textwidth]{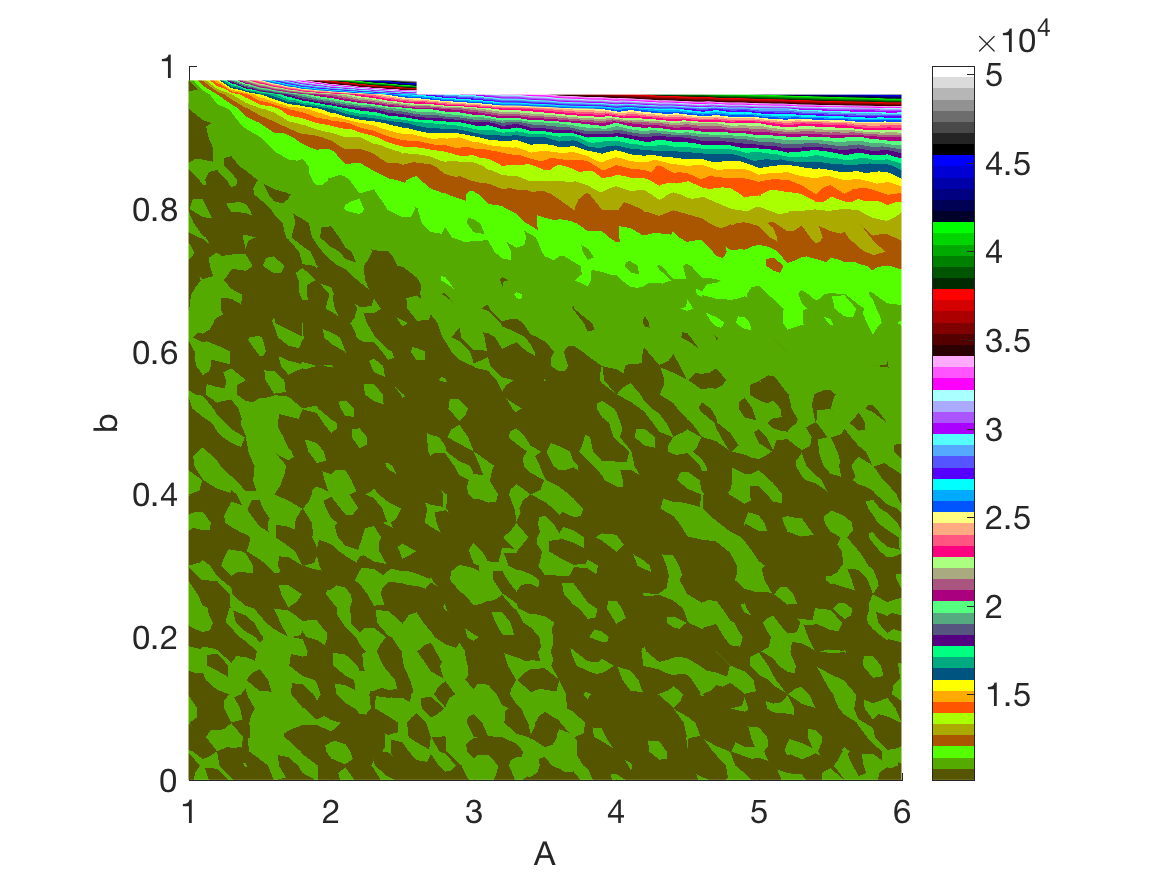}
                \subcaption{\onemax with $n=1500$}\label{fig:OM1500}
        \end{minipage}\hfill
				\begin{minipage}[t]{0.33\textwidth}
                \includegraphics[trim={12mm 0 2mm 0},width=\textwidth]{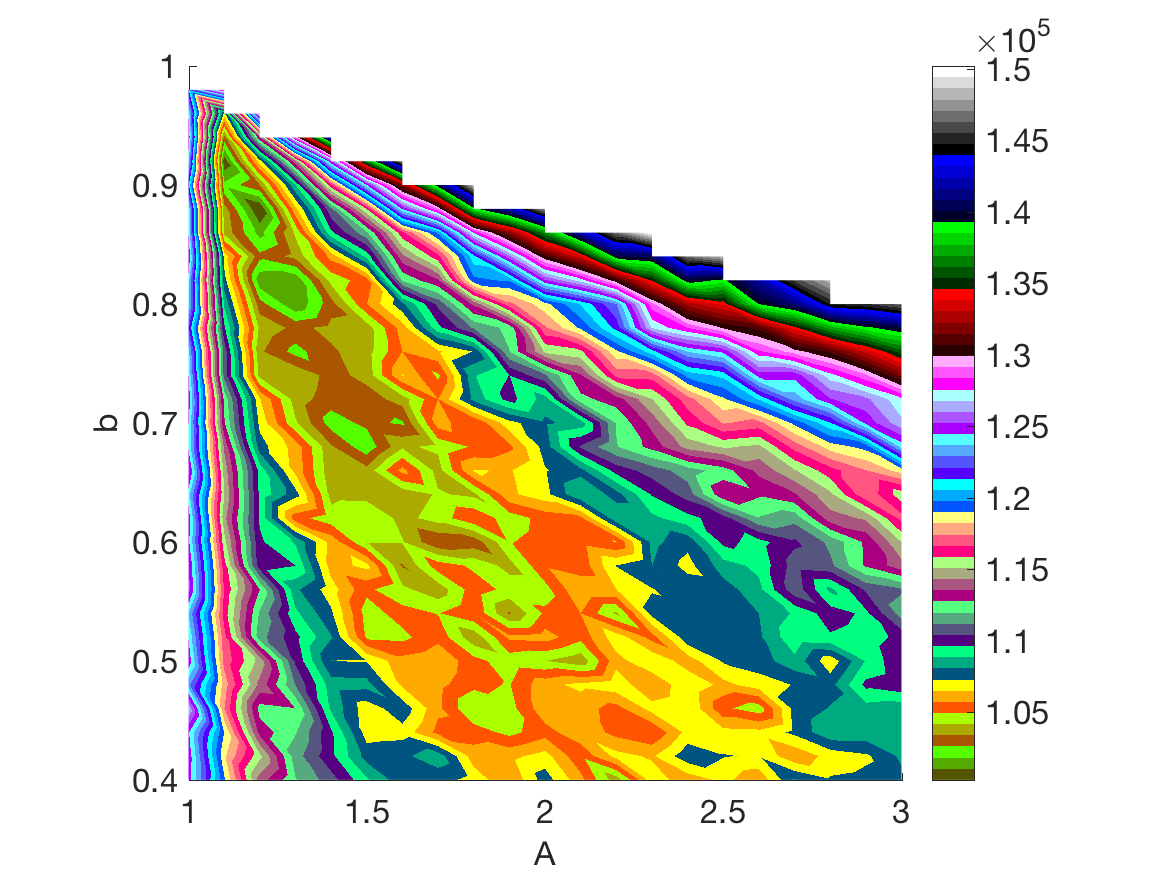}
                \subcaption{\leadingones with $n=500$}\label{fig:LO500-zoom}
        \end{minipage}
				      \medskip
				
        \begin{minipage}[t]{0.33\textwidth}
                \includegraphics[trim={12mm 0 2mm 0},width=\textwidth]{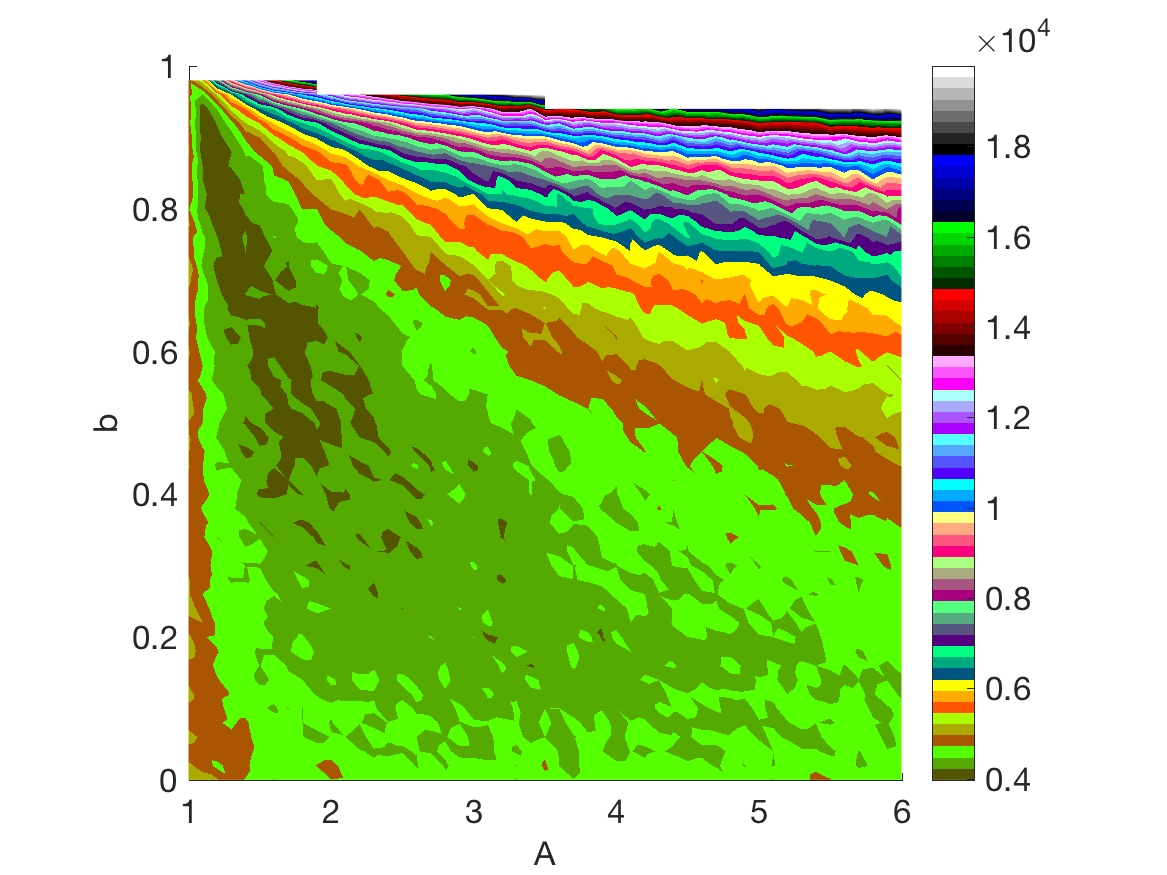}
                \subcaption{\leadingones with $n=100$}\label{fig:LO100}
        \end{minipage}\hfill
        \begin{minipage}[t]{0.33\textwidth}
                \includegraphics[trim={12mm 0 2mm 0},width=\textwidth]{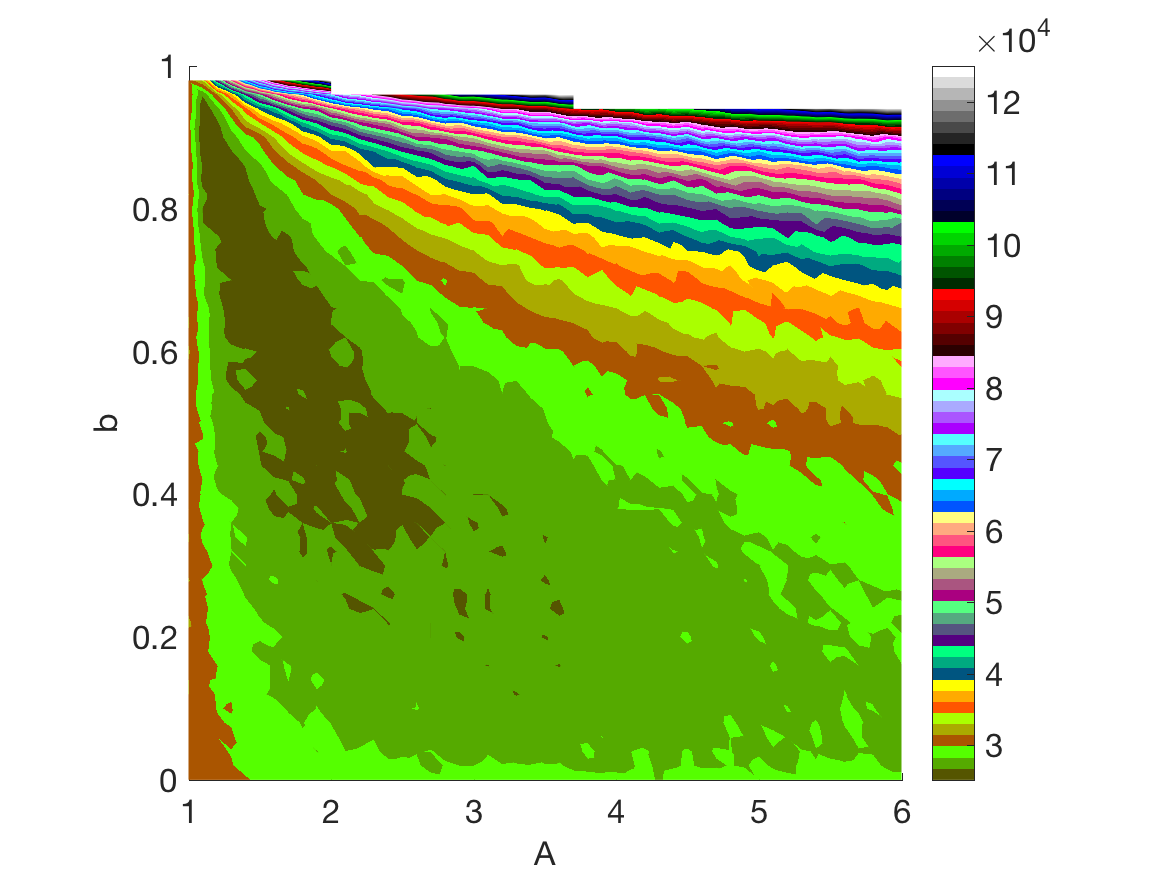}
                \subcaption{\leadingones with $n=250$}\label{fig:LO250}
        \end{minipage}
				 \begin{minipage}[t]{0.33\textwidth}
                \includegraphics[trim={12mm 0 2mm 0},width=\textwidth]{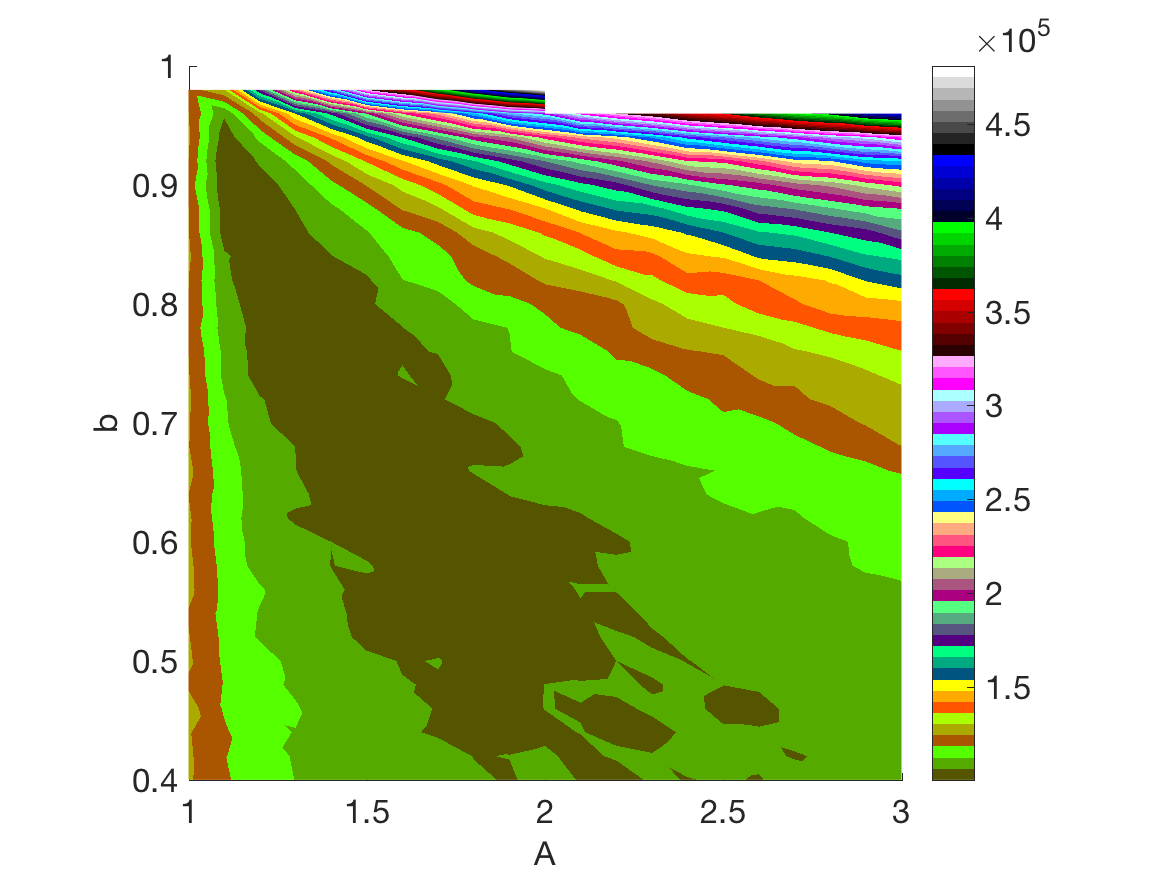}
                \subcaption{\leadingones with $n=500$}\label{fig:LO500}
        \end{minipage}

				\caption{Optimization times of the \oeaalpha, averaged over 101 independent runs}
				\label{fig:grid}
\end{figure*}

With the description of the algorithms and benchmark problems in place, we can now report our empirical results for the \oeaalpha. In a first step, we are interested in the sensitivity of the \oeaalpha with respect to the two update strengths $A$ and $b$. To analyze the influence of these two \emph{hyper-parameters}, we perform a grid search, in which we run the \oeaalpha for various combinations of $A$ and $b$. For these experiments, we always initialize the mutation rate as $p_0=1/n$. In Figures~\ref{fig:OM500} to~\ref{fig:LO500} we report for each configuration the average optimization times of 101 independent runs of these algorithms on \onemax (Figures~\ref{fig:OM500} and~\ref{fig:OM1500}) and on \leadingones (Figures~\ref{fig:LO100}, \ref{fig:LO250}, \ref{fig:LO500}). In these plots, the parameters are chosen as follows. For \onemax and for $\leadingones$ with $n\le 250$, we vary $A$ between $1.0$ and $6.0$, in multiples of $0.1$, and we choose $b$ between $0.00$ and $1.00$, in multiples of 0.02. For \leadingones with $n=500$ we restrict the values to $1 \leq A \leq 3$ and $0.4 \leq b \leq 1$. That is, Figure~\ref{fig:LO500} is a zoom into the upper left region of the full grid search.

These graphics are to be read as follows. As so-called \emph{heat maps}, we visualize in color the optimization times of the \oeaalpha variants; that is, we plot the average number of fitness evaluations that these algorithms needed in order to locate the global optimum. We use a binning of colors to emphasize the visibility of regions and gradients. For example, the large green regions indicate the configurations for which the \oeaalpha performs best. Also, the quick succession of colors in the top right corner shows that, beyond some threshold values for $A$ and $b$, small changes in the configuration can cause large changes in the performance. 

For \onemax we first observe that the heat maps have a very similar structure across the different dimensions, cf. Figures~\ref{fig:OM500} and~\ref{fig:OM1500}. In addition to the results shown in Figure~\ref{fig:grid} we also computed heat maps for \onemax with problem dimension $n=100$, $n=500$, and $n=1,500$ and for each of these heat maps the overall structure is very similar to those plotted in Figure~\ref{fig:grid}. Also for \leadingones the figures are quite similar across the dimensions, cf. Figures~\ref{fig:LO100}, \ref{fig:LO250}, and~\ref{fig:LO500} (recall that Figure~\ref{fig:LO500} is a zoom into upper left corner). We also observe that the performance landscapes for \onemax is quite flat; i.e., the bulk of the configurations achieves a very similar performance. 

For both problems, \onemax and \leadingones, we see that for large values of $A$ and $b$ the average optimization times become worse very quickly. In Figure~\ref{fig:LO500-zoom} we therefore zoom into the most interesting region of high-performing configurations and display only results for configurations that achieve an average optimization time that is at most $150,000$ (for comparison, the \oeares needs around $135,700$ iterations, on average, on this problem instance, and RLS $125,000$, cf. Table~\ref{tab:LOn}). This zoom increases the granularity of the color scheme, and allows to detect more structure within this region.

As a next step, we compare the average running times with those of the \oeares and RLS. We observe that a significant number of configurations outperform the static \oeares. For \leadingones, the average optimization times of different algorithms can be found in Table~\ref{tab:LOn} and for \onemax we note that the average optimization time of the \oeares is around $4,750$ for $n=500$ and about $16,630$ for $n=1,500$. Several configurations also outperform RLS, but for \onemax the 101 runs do not suffice to make a statistically sound comparison, since the advantage of adaptive step sizes is bounded by around $2\%$ for all tested dimensions. For \leadingones, however, the advantages over RLS are quite significant, as Figure~\ref{fig:Comparison-LO-BetterRLS} demonstrates. In this plot, the lowermost two lines illustrate the fraction of all 2,450 configurations with $1<A\le 6$ and $0< b <1$ that yield a better average performance on \leadingones than RLS. In this figure, the $x$-axis indicates the percentage by which the algorithms are better than RLS, and on the $y$-axis we plot the fraction of the configurations that outperform RLS by at least this much. That is, we see that among all 2,450 configurations around $67\%$ have an average optimization time below $n^2/2$. Between $37\%$ ($n=100$) and $41\%$ (n=250) of all configurations are better than RLS by at least $10\%$. 

When we restrict the configurations to those 450 that satisfy $1<A\le 2.5$ and $0.4\le b <1$ (three uppermost lines), around $78\%$ of them are better than RLS, around $62\%$ are better by at least $10\%$, around $30\%$ excel over RLS by at least $15\%$, and almost $20\%$ of the configurations are better than $16\%$. From an algorithm design point of view this is very good news: finding good hyper-parameters is not very difficult for this problem. We also see that the numbers are very similar across all three tested dimensions $n=100,250,500$, suggesting that this surprisingly good performance might translate to larger dimensions.

As we have proven in Section~\ref{sec:theory}, the best possible running time on \leadingones (among all unary unbiased black-box algorithms, and hence in particular among all \oea and RLS variants) is achieved by $\RLS_{\text{opt,LO}}$. For the considered problem dimensions, $\RLS_{\text{opt,LO}}$ is better, in expectation, than RLS by around $22.3\%$, cf. Table~\ref{tab:LOn}. This advantage over RLS is of course also the maximal improvement that any \oeaalpha variant can achieve over RLS. For all dimensions, we observe that between $2$ and $4\%$ of the configurations are better than RLS by at least $18\%$. No configuration achieves a $20\%$ improvement. Note here that this does not come as a surprise: even if the \oea chooses in each iteration the for this state optimal mutation rate, its performance still suffers from the random choice of the step sizes. This risk is eliminated in RLS and \RLSopt by the deterministic choice of the mutation strength $\ell=1$ and $\ell=k_{\text{opt,\LO}}(n,\LO(x))$, respectively. We can therefore not expect any configuration, or, more generally, any evolutionary algorithm, to achieve the same performance as $\RLS_{\text{opt,LO}}$. 

\begin{figure}
\begin{center}
\includegraphics[width=0.7\linewidth]{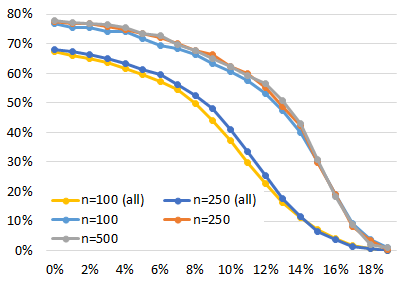}
\end{center}
\caption{Fraction of all 2,450 (450) configurations with $1<A\le 6$ and $0<b<1$  ($1<A\le 2.5$ and $0.4\le b<1$) that outperform RLS on \leadingones by at least $x\%$}
\label{fig:Comparison-LO-BetterRLS}
\end{figure}

In comparison to the \oeares, the comparison is even more impressive. About $73\%$ of all $2,450$ and around $82\%$ of the $450$ restricted configurations with $1<A\le 2.5$ and $0.4\le b <1$ have a better performance on \leadingones than the \oeares. About $64\%$ ($76\%$ for the restricted hyper-parameters) are better by at least $10\%$ and still around $14\%$ ($47\%$) are better, on average, by more than $20\%$. The best improvements over the \oeares are around $26\%$, but we should keep in mind here that the numbers are averages for 101 independent runs only. We will therefore do a more thorough investigation of selected configurations in the next subsection. Despite the variance of the algorithmic performance, however, we see a quite consistent behavior in Figure~\ref{fig:Comparison-LO-BetterRLS}, raising our confidence that these results are not much skewed by the relatively low number of independent runs. 

For \onemax, around $90\%$ of all configurations perform better than the \oeares, and between $64\%$ ($n=100$) and $83\%$ ($n=2,000$) are better by at least $30\%$. The best configurations achieve an improvement of up to around $40\%$. 

\section{Detailed Results for LeadingOnes}~\label{sec:experimentsLO}

To substantiate the comparisons made in Section~\ref{sec:grid} and to investigate how the results translate to larger problem dimensions, we now take a closer look at some selected configurations. In this section we regard \leadingones. Detailed results for \onemax can be found in Section~\ref{sec:experimentsOM} below.

For \leadingones, the results of this comparison are summarized in Table~\ref{tab:LOn}, where we report both the average optimization times $T$ and the relative values $T/n^2$. For RLS and $\RLS_{\text{opt,LO}}$ the numbers in Table~\ref{tab:LOn} are computed from the exact running time statements (cf. Section~\ref{sec:theory}), which are all indeed very close to the results that we obtain empirically. All other numbers in Table~\ref{tab:LOn} are averages over $1,001$ and $101$ independent runs, respectively.

In addition to the \oeares, we list 4 different configurations. The \oeaalpha with $A=1.3, b=0.75$ and with $A=1.2, b=0.85$ are examples for configurations that show a good (but not empirically best) performance in the grid search conducted in Section~\ref{sec:grid}. We also add to the comparison the configuration $A=2,b=0.5$, a seemingly intuitive configuration used also in different context, e.g., in the adaptive choice of the population size in~\cite{LassigS11}. The configuration $A=1.11, b=0.66$ corresponds to the $1/5$-th rule with update strength $1.5$, a very common adaptation rule in continuous domain (cf.~\cite{DoerrD15self} for a discussion and references).

We observe that the performances are quite stable over the tested dimensions. The two selected configurations $A=1.3, b=0.75$ and $A=1.2, b=0.85$ are better by around $18\%$ than RLS, on all tested dimensions. The doubling/halving rule $A=2,b=0.5$ achieves a $15-17\%$ performance gain over RLS, while the 1/5-th rule with $A=1.11, b=0.66$ achieves an improvement over RLS of about $7-10\%$.

As an important consequence, the results of this section suggest that a tuning of the hyper-parameters $A$ and $b$ on smaller dimension is possible. In addition, the stability of the results indicates that similar results as those presented in Section~\ref{sec:grid} are likely to apply also to larger problem dimensions.

\setlength{\tabcolsep}{.5mm}\renewcommand{\arraystretch}{1.2}\begin{table*}\begin{center}{\scalebox{0.93}{\footnotesize{\begin{tabular}{c|c|c|c|c|c|c|c|c|c|c}\hline
	&	\multicolumn{2}{c|}{$n=100$}			& 	\multicolumn{2}{c|}{$n=250$}			& 	\multicolumn{2}{c|}{$n=500$}			& 	\multicolumn{2}{c|}{$n=1000$}			& 	\multicolumn{2}{c}{$n=1500$}			\\
Algorithm 	& 	$T$	& 	$T/n^2$	& 	$T$	& 	$T/n^2$	& 	$T$	& 	$T/n^2$	& 	$T$	& 	$T/n^2$	& 	$T$	& 	$T/n^2$	\\
\hline																					
$\RLS_{\text{opt,LO}}$	& 	 3,883   	& 	38.83\%	& 	 24,273   	& 	38.8368\%	& 	 97,102   	& 	38.8408\%	& 	 388,427   	& 	38.8427\%	& 	 873,981   	& 	38.8436\%	\\
RLS	& 	 5,001   	& 	50.01\%	& 	 31,251   	& 	50.0016\%	& 	 125,001   	& 	50.0004\%	& 	 500,001   	& 	50.0001\%	& 	 1,125,001   	& 	50.0000\%	\\
\hline
(1+1) EA$_{>0}$	& 	 5,401   	& 	54.01\%	& 	 33,817   	& 	54.1072\%	& 	 135,782   	& 	54.3128\%	& 	 544,288   	& 	54.4288\%	& 	 1,216,448*   	& 	54.0644\%	\\
\hline																					
(1+1) EA$_\alpha$(A=1.2, b=0.85)	& 	 4,063   	& 	40.63\%	& 	 25,497   	& 	40.7952\%	& 	 101,976   	& 	40.7904\%	& 	 409,820*   	& 	40.9820\%	& 	 921,900*   	& 	40.9733\%	\\
(1+1) EA$_\alpha$(A=1.3, b=0.75)	& 	 4,185   	& 	41.85\%	& 	 25,731   	& 	41.1696\%	& 	 102,985   	& 	41.1940\%	& 	 413,518*   	& 	41.3518\%	& 	 931,313*   	& 	41.3917\%	\\
\hline																					
(1+1) EA$_\alpha$(A=2.0, b=0.5)	& 	 4,195   	& 	41.95\%	& 	 26,247   	& 	41.9952\%	& 	 104,193*   	& 	41.6772\%	& 	416,362*	& 41.6362\%		& 	932,791*	& 	41.4574\%	\\
(1+1) EA$_\alpha$(A=1.11, b=0.66)	& 	 4,495   	& 	44.95\%	& 	 28,277	& 	45,2432\%	& 	 114,814*   	& 	45.9256\%	& 	448,815*   	& 	44.8815\%	& 	1,016,830*   	& 	45.1924\%		\\
\hline
\end{tabular}
}}}\end{center}
\caption{Average optimization times for \leadingones. Exact bounds for RLS and $\RLS_{\text{opt,LO}}$, empirical averages over 1,001 (*=101) independent runs otherwise.}
\label{tab:LOn}
\end{table*}

\subsection{Zooming into Typical Runs}\label{sec:zoomLO}

We finally want to understand how well the selected mutation rates resemble the optimal ones. Our benchmark problem is the 500-dimensional \leadingones function. We store for 10 independent runs of the \oeaalpha with $A=1.2$ and $b=0.85$ and for each iteration the $\LO$-value of a best-so-far solution along with the mutation strength $\ell$ that has been chosen in this iteration, i.e., the number of bits that have been flipped by the $\mut_{\ell}$ operator to create the offspring of this iteration.  We then average for each function value $\LO(x)$ over the $\ell$-values that have been used in iterations that started with this fitness value. Figure~\ref{fig:ell-LO} plots these averages for $3\le \LO(x) < 100$ (blue, ragged curve), along with the corresponding $k_{\text{opt,LO}}(n,\LO(x))$ values (smooth red curve). The \oeaalpha seems to sample indeed almost optimal mutation strengths. For $\LO(x) \in \{0,1,2\}$ we note that the values are smaller than $k_{\text{opt,LO}}(n,\LO(x))$, but this is explained by the initialization of the mutation rate with $p_0=1/n$, which forces the algorithm to first increase this rate to a close-to-optimal value. This process decreases the average considerably. For values $\LO(x) \ge 100$, the two curves are almost indistinguishable and have therefore been removed from the illustration. 
\begin{figure}
\begin{center}
\includegraphics[width=0.7\linewidth]{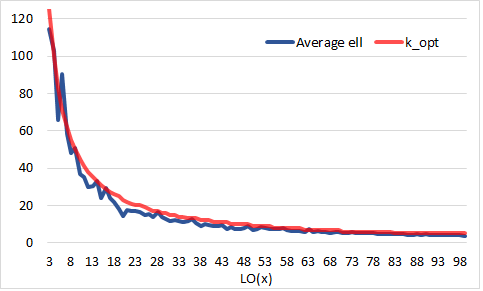}
\end{center}
\caption{Average and optimal mutation strengths for different $\LO(x)$ values ($n=500$, 10 independent runs of the \oeaalpha with $A=1.2$, $b=0.85$, and $p_0=1/n$)}
\label{fig:ell-LO}
\end{figure}

\section{Detailed Results for OneMax}~\label{sec:experimentsOM}

Similar to Figure~\ref{fig:Comparison-LO-BetterRLS} we display in Figure~\ref{fig:fractionOM} the fraction of all 2,450 (600 for $n=2,000$) tested configurations with $1<A\le 6$ and $0<b<1$ that are better than the \oeares. Note that, for \onemax, we cannot expect the configurations to outperform RLS, since the performance of RLS and \RLSopt are very close: for the here-tested problem dimensions the difference between these two algorithms is only around $2\%$. In light of the high variance of the running times, statistically sound comparison would require a lot of independent repetitions. We have to leave this aspect for future work. 

\begin{figure}[h!]
\begin{center}
\includegraphics[width=.5\linewidth]{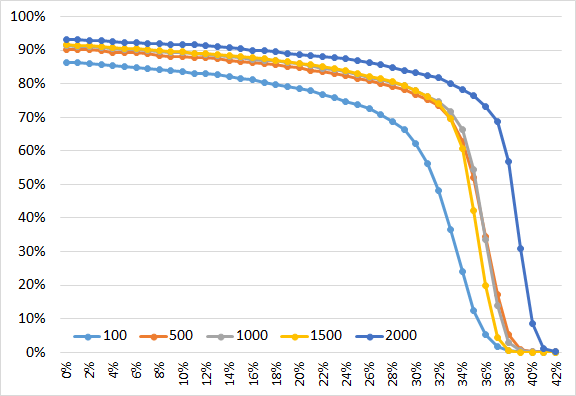}
\end{center}
\caption{Fraction of all 2,450 (600 for $n=2,000$) configurations with $1<A\le 6$ and $0<b<1$  ($1<A\le 3.0$ and $0.4\le b<1$) that outperform the \oeares on \onemax by at least $x\%$}
\label{fig:fractionOM}
\end{figure}

We also note that our main objective with the \onemax benchmark is to understand if the \oeaalpha is capable of identifying that the optimal mutation strength very quickly converges to $1$ as the optimization proceeds. This aspect is considered in Figure~\ref{fig:ell-OM}, where we plot for the 2000-dimensional \onemax function and the configuration with $A=1.2$ and $b=0.85$ for 11 independent runs the average value of $\ell$ that has been chosen at a given function value. We zoom in this picture into the most interesting region of \OM-values between $1,050$ and $1,360$. Just as in Figure~\ref{fig:ell-LO} we observe that the $\ell$-values first have to be increased, since $p_0$ is initialized as $1/n$, whereas the optimal mutation rate is around $1/2$ for $\OM(x)=1,000=n/2$ (recall that the random initial solution has a function value of around $1,000$, so that we do not regard $\OM(x)$ values below this value). While the initial values are certainly smaller than $k_{\text{opt,OM}}(n,\OM(x))$ for $\OM(x)<1,230$, we see that they are still much above the static choice $\ell=1$ used by RLS. For \OM-values greater than $1,332$ the one-bit flips are optimal, i.e., $k_{\text{opt,OM}}(n,\OM(x))=1$ whenever $\OM(x)> 1,332$. All average $\ell$-values are very close to one in this regime. 

\begin{figure}[h!]
\centering
\begin{center}
\includegraphics[width=0.5\linewidth]{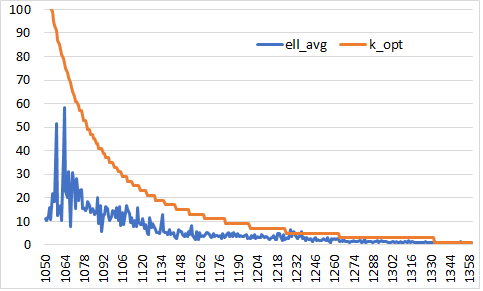}
\end{center}
\caption{Average and optimal mutation strengths for different $\OM(x)$ values ($n=2,000$, 11 independent runs of the \oeaalpha with $A=1.2$, $b=0.85$, and $p_0=1/n$)}
\label{fig:ell-OM}
\end{figure}

We also conducted experiments for the same configurations as in Table~\ref{tab:LOn}, for problem dimensions up to $n=3,000$. The results are similar to those for \leadingones in that also for \onemax the relative performance gains observed in small dimensions seem to transfer to larger ones. It is remarkable also that all the selected configurations, on average over $1,001$ independent runs, achieve a performance that is very close to that of RLS, and in some instances even outperform it. The following table summarizes selected results. For convenience, we also compute for each empirical average $T$ the value $c:=T/(n \ln n)$. Starred results are for 101 independent runs, all other results are averages over $1,001$ runs. 

\setlength{\tabcolsep}{1.5mm}\renewcommand{\arraystretch}{1.2}\begin{table*}[h!]
\begin{center}
\scalebox{0.87}{
\begin{tabular}{c|cc|cc|cc|cc|cc}
\hline
	&	\multicolumn{2}{c|}{$n=100$}			& 	\multicolumn{2}{c|}{$n=500$}			& 	\multicolumn{2}{c|}{$n=1000$}			& 	\multicolumn{2}{c|}{$n=2000$}			& 	\multicolumn{2}{c}{$n=3000$}			\\
Algorithm 	& 	$T$	& 	$c$	& 	$T$	& 	$c$	& 	$T$	& 	$c$	& 	$T$	& 	$c$	& 	$T$	& 	$c$	\\
\hline																					
RLS	& 	 445   	& 	 0.9663   	& 	 3,050   	& 	 0.9816   	& 	 6,871   	& 	 0.9947   	& 	 14,809   	& 	 0.9742   	& 	 23,814   	& 	 0.9915   	\\
RLS$_{\opt}$	& 	 436   	& 	 0.9474   	& 	 2,974   	& 	 0.9572   	& 	 6,690   	& 	 0.9685   	& 	 14,722   	& 	 0.9684   	& 	 23,507   	& 	 0.9787   	\\
\hline																					
(1+1) EA$_{>0}$	& 	 679   	& 	 1.4738   	& 	 4,756   	& 	 1.5306   	& 	 10,574   	& 	 1.5307   	& 	 24,352   	& 	 1.6019   	& 	 37,256   	& 	 1.5511   	\\
\hline																					
(1+1) EA$_\alpha$(A=1.2, b=0.85)	& 	 450   	& 	 0.9776   	& 	 3,059   	& 	 0.9845   	& 	 6,751   	& 	 0.9773   	& 	 14,801   	& 	 0.9736   	& 	 23,558   	& 	 0.9808   	\\
(1+1) EA$_\alpha$(A=1.3, b=0.75)	& 	 450   	& 	 0.9767   	& 	 3,033   	& 	 0.9761   	& 	 6,801   	& 	 0.9845   	& 	 14,974   	& 	 0.9850   	& 	 23,715   	& 	 0.9873   	\\
\hline																					
(1+1) EA$_\alpha$(A=2.0, b=0.5)	& 	 455   	& 	 0.9872   	& 	 3,013   	& 	 0.9697   	& 	 6,753   	& 	 0.9776   	& 	14613	& 	 0.9613   	& 	 23,027   	& 	 0.9587   	\\
(1+1) EA$_\alpha$(A=1.11, b=0.66)	& 	 447   	& 	 0.9704   	& 	 3,039   	& 	 0.9780   	& 	 6,749   	& 	 0.9770   	& 	 15,134   	& 	 0.9955   	& 	 24,011   	& 	 0.9997   	\\
\hline
\end{tabular}}
\end{center}
\caption{Average optimization times for \onemax for 1,001 (*=101) independent runs. $c:=100T/(n \ln n)$}
\label{tab:OMn}
\end{table*}

\section{{Conclusions}\label{sec:conclusions}}

We hope to contribute with our work to a more widespread experimentation and use of non-static parameter selection mechanisms in discrete optimization contexts. We are confident that significant performance gains are possible, for a broad number of applications. We have shown in this work that already quite simple parameter control mechanisms can give almost optimal performance. On \leadingones, significant performance gains over the best static parameter values were possible for a broad range of multiplicative update rules. 

Much more sophisticated parameter control techniques, including a number of portfolio-based methods inspired by the multi-armed bandit literature have been proposed and analyzed in the literature, by experimental~\cite{Thierens05,FialhoCSS08,FialhoCSS09} and theoretical~\cite{DoerrDY16PPSN,LissovoiOW17} means. Developing a rigorous understanding of which update scheme to favor under which circumstances is the ultimate goal of our research. First comparisons with the above-mentioned techniques are quite favorable for the success-based multiplicative update rule; a rigorous comparison is left for future work. 

As a more immediate research question, it would be desirable to understand how the performance of the \oeaalpha scales with very large dimensions, by means of an empirical comparison and/or a mathematical running time analysis. 

\subsection*{Acknowledgments}

The authors would like to thank Eduardo Carvalho Pinto for providing his implementation of the \oeaalpha and his contributions to a preliminary experimentation with the multiplicative parameter control mechanism.

Our work was supported by a public grant as part of the
Investissement d'avenir project, reference ANR-11-LABX-0056-LMH,
LabEx LMH and by the Australian Research Council project DE160100850.


\begin{thebibliography}{DGWY17}

\bibitem[BDN10]{BottcherDN10}
S\"untje B{\"o}ttcher, Benjamin Doerr, and Frank Neumann.
\newblock Optimal fixed and adaptive mutation rates for the {L}eading{O}nes
  problem.
\newblock In {\em Proc. of Parallel Problem Solving from Nature (PPSN'10)},
  volume 6238 of {\em Lecture Notes in Computer Science}, pages 1--10.
  Springer, 2010.

\bibitem[DD15]{DoerrD15self}
Benjamin Doerr and Carola Doerr.
\newblock Optimal parameter choices through self-adjustment: Applying the
  1/5-th rule in discrete settings.
\newblock In {\em Proc. of Genetic and Evolutionary Computation Conference
  (GECCO'15)}, pages 1335--1342. {ACM}, 2015.

\bibitem[DD16]{DoerrD16impact}
Benjamin Doerr and Carola Doerr.
\newblock The impact of random initialization on the runtime of randomized
  search heuristics.
\newblock {\em Algorithmica}, 75:529--553, 2016.

\bibitem[DDK16]{DoerrDK16PPSN}
Benjamin Doerr, Carola Doerr, and Timo K{\"{o}}tzing.
\newblock Provably optimal self-adjusting step sizes for multi-valued decision
  variables.
\newblock In {\em Proc. of Parallel Problem Solving from Nature (PPSN'16)},
  volume 9921 of {\em LNCS}, pages 782--791. Springer, 2016.

\bibitem[DDY16a]{DoerrDY16PPSN}
Benjamin Doerr, Carola Doerr, and Jing Yang.
\newblock $k$-bit mutation with self-adjusting $k$ outperforms standard bit
  mutation.
\newblock In {\em Proc. of Parallel Problem Solving from Nature (PPSN'16)},
  volume 9921 of {\em LNCS}, pages 824--834. Springer, 2016.

\bibitem[DDY16b]{DoerrDY16}
Benjamin Doerr, Carola Doerr, and Jing Yang.
\newblock Optimal parameter choices via precise black-box analysis.
\newblock In {\em Proc. of Genetic and Evolutionary Computation Conference
  (GECCO'16)}, pages 1123--1130. {ACM}, 2016.

\bibitem[DGWY17]{DoerrGWY17}
Benjamin Doerr, Christian Gie{\ss}en, Carsten Witt, and Jing Yang.
\newblock The $(1+\lambda)$~evolutionary algorithm with self-adjusting mutation
  rate.
\newblock In {\em Proc. of Genetic and Evolutionary Computation Conference
  (GECCO'17)}, pages 1351--1358. ACM, 2017.

\bibitem[DL16a]{DangL16PPSN}
Duc-Cuong Dang and Per~Kristian Lehre.
\newblock Self-adaptation of mutation rates in non-elitist populations.
\newblock In {\em Proc. of Parallel Problem Solving from Nature (PPSN'16)},
  volume 9921 of {\em LNCS}, pages 803--813. Springer, 2016.

\bibitem[DL16b]{DoerrL16}
Carola Doerr and Johannes Lengler.
\newblock The {(1+1)} elitist black-box complexity of {L}eading{O}nes.
\newblock In {\em Proc. of Genetic and Evolutionary Computation Conference
  (GECCO'16)}, pages 1131--1138. ACM, 2016.

\bibitem[Doe17]{Carola17}
Carola Doerr.
\newblock Non-static parameter choices in evolutionary computation.
\newblock In {\em Companion Material for Proc. of Genetic and Evolutionary
  Computation Conference (GECCO'17)}, pages 736--761. ACM, 2017.

\bibitem[FCSS08]{FialhoCSS08}
{\'{A}}lvaro Fialho, Lu{\'{\i}}s~Da Costa, Marc Schoenauer, and Mich{\`{e}}le
  Sebag.
\newblock Extreme value based adaptive operator selection.
\newblock In {\em Proc. of Parallel Problem Solving from Nature (PPSN'08)},
  volume 5199 of {\em LNCS}, pages 175--184. Springer, 2008.

\bibitem[FCSS09]{FialhoCSS09}
{\'{A}}lvaro Fialho, Lu{\'{\i}}s~Da Costa, Marc Schoenauer, and Mich{\`{e}}le
  Sebag.
\newblock Dynamic multi-armed bandits and extreme value-based rewards for
  adaptive operator selection in evolutionary algorithms.
\newblock In {\em Proc. of Learning and Intelligent Optimization (LION'09)},
  volume 5851 of {\em LNCS}, pages 176--190. Springer, 2009.

\bibitem[FCSS10]{FialhoCSS10}
{\'{A}}lvaro Fialho, Lu{\'{\i}}s~Da Costa, Marc Schoenauer, and Mich{\`{e}}le
  Sebag.
\newblock Analyzing bandit-based adaptive operator selection mechanisms.
\newblock {\em Annals of Mathematics and Artificial Intelligence}, 60:25--64,
  2010.

\bibitem[JDW05]{JansenJW05}
Thomas Jansen, Kenneth~A. {De Jong}, and Ingo Wegener.
\newblock On the choice of the offspring population size in evolutionary
  algorithms.
\newblock {\em Evolutionary Computation}, 13:413--440, 2005.

\bibitem[JZ11]{JansenZ11}
Thomas Jansen and Christine Zarges.
\newblock Analysis of evolutionary algorithms: from computational complexity
  analysis to algorithm engineering.
\newblock In {\em Proc. of Foundations of Genetic Algorithms (FOGA'11)}, pages
  1--14. {ACM}, 2011.

\bibitem[JZ14]{JansenZ14}
Thomas Jansen and Christine Zarges.
\newblock Performance analysis of randomised search heuristics operating with a
  fixed budget.
\newblock {\em Theoretical Computer Science}, 545:39--58, 2014.

\bibitem[KGV83]{SA83}
Scott Kirkpatrick, C.~D. Gelatt, and Mario~P. Vecchi.
\newblock Optimization by simulated annealing.
\newblock {\em Science}, 220(4598):671--680, 1983.

\bibitem[KHE15]{KarafotiasHE15}
G.~Karafotias, M.~Hoogendoorn, and A.E. Eiben.
\newblock Parameter control in evolutionary algorithms: Trends and challenges.
\newblock {\em IEEE Transactions on Evolutionary Computation}, 19:167--187,
  2015.

\bibitem[LOW17]{LissovoiOW17}
Andrei Lissovoi, Pietro~Simone Oliveto, and John~Alasdair Warwicker.
\newblock On the runtime analysis of generalised selection hyper-heuristics for
  pseudo-boolean optimisation.
\newblock In {\em Proc. of Genetic and Evolutionary Computation Conference
  (GECCO'17)}, pages 849--856. ACM, 2017.

\bibitem[LS11]{LassigS11}
J{\"o}rg L{\"a}ssig and Dirk Sudholt.
\newblock Adaptive population models for offspring populations and parallel
  evolutionary algorithms.
\newblock In {\em Proc. of Foundations of Genetic Algorithms (FOGA'11)}, pages
  181--192. ACM, 2011.

\bibitem[LW12]{LehreW12}
Per~Kristian Lehre and Carsten Witt.
\newblock Black-box search by unbiased variation.
\newblock {\em Algorithmica}, 64:623--642, 2012.

\bibitem[PD17]{CarvalhoD17}
Eduardo~Carvalho Pinto and Carola Doerr.
\newblock Discussion of a more practice-aware runtime analysis for evolutionary
  algorithms.
\newblock In {\em Proc. of Artificial Evolution (EA'17)}, pages 298--305, 2017.

\bibitem[Thi05]{Thierens05}
Dirk Thierens.
\newblock An adaptive pursuit strategy for allocating operator probabilities.
\newblock In {\em Proc. of Genetic and Evolutionary Computation Conference
  (GECCO'05)}, pages 1539--1546. ACM, 2005.

\bibitem[Thi09]{Thierens09}
Dirk Thierens.
\newblock On benchmark properties for adaptive operator selection.
\newblock In {\em Proc. of Genetic and Evolutionary Computation Conference
  (GECCO'09), Companion Material}, pages 2217--2218. ACM, 2009.

\bibitem[vHB02]{HemertB02}
Jano~I. van Hemert and Thomas B{\"{a}}ck.
\newblock Measuring the searched space to guide efficiency: The principle and
  evidence on constraint satisfaction.
\newblock In {\em Proc. of Parallel Problem Solving from Nature (PPSN'02)},
  volume 2439 of {\em LNCS}, pages 23--32. Springer, 2002.

\end{thebibliography}

\end{document}